\documentclass{bmvc2k}

\usepackage{booktabs}
\usepackage{adjustbox}
\usepackage{multirow}

\usepackage{amssymb}

\usepackage{algorithm}
\usepackage{algorithmic}
\usepackage{float}

\title{CLFSeg: A Fuzzy-Logic based Solution for Boundary Clarity and Uncertainty Reduction in Medical Image Segmentation}

\addauthor{Anshul Kaushal\textsuperscript{*, $\dagger$, }}{kaushalanshul.itz@gmail.com}{1}
\addauthor{Kunal Jangid\textsuperscript{*,}}{kunal24@iiserb.ac.in}{2}
\addauthor{Vinod K. Kurmi}{vinodkk@iiserb.ac.in}{2}

\addinstitution{
     UIET-H, Panjab University, \\ 
 Hoshiarpur, India
}

\addinstitution{
 Department of Data Science and Engineering \\
 Indian Institute of Science Education and Research\\
 Bhopal, India
}

\runninghead{A.Kaushal, K.Jangid, V.K.Kurmi}{CLFSeg: Fuzzy-Logic Segmentation}

\begin{document}

\maketitle

\begingroup
\renewcommand\thefootnote{\fnsymbol{footnote}}
\footnotetext[1]{Equal contribution.}
\renewcommand\thefootnote{$\dagger$}
\footnotetext{Work done at IISER Bhopal.}
\endgroup

\begin{abstract}
Accurate polyp and cardiac segmentation for early detection and treatment is essential for the diagnosis and treatment planning of cancer-like diseases. Traditional convolutional neural network (CNN) based models have represented limited generalizability, robustness, and inability to handle uncertainty, which affects the segmentation performance. To solve these problems, this paper introduces CLFSeg, an encoder-decoder based framework that aggregates the Fuzzy-Convolutional (FC) module leveraging convolutional layers and fuzzy logic. This module enhances the segmentation performance by identifying local and global features while minimizing the uncertainty, noise, and ambiguity in boundary regions, ensuring computing efficiency. 
In order to handle class imbalance problem while focusing on the areas of interest with tiny and boundary regions, binary cross-entropy (BCE) with dice loss is incorporated. Our proposed model exhibits exceptional performance on four publicly available datasets, including CVC-ColonDB, CVC-ClinicDB, EtisLaribPolypDB, and ACDC.
Extensive experiments and visual studies show CLFSeg surpasses the existing SOTA performance and focuses on relevant regions of interest in anatomical structures. The proposed CLFSeg improves performance while ensuring computing efficiency, which makes it a potential solution for real-world medical diagnostic scenarios. Project page is available at \url{https://visdomlab.github.io/CLFSeg/}.
\end{abstract}

\section{Introduction}
\label{sec:intro}

The advent of deep learning algorithms has significantly impacted biomedical research, particularly in polyp and cardiac image segmentation. 
Polyps are growths of abnormal tissue in the colon that can be signs of colorectal cancer. It is important to accurately segment polyps to lower the death rate from cancer.
So, polyp segmentation is a vital problem in medical imaging, especially for the early identification and prevention of colorectal cancer.
The manual analysis of these images can be labor-intensive and prone to inaccuracies due to  intricate borders, and diversity of polyps in size, shape, appearance, and low-contrast areas.

Traditional segmentation methods like convolutional neural networks (CNNs) are time-consuming and lack precision. 
 Recent advances have shown promising results, especially deep learning-based architectures such as ResNet \cite{he2016deep}, U-Net \cite{ronneberger2015u}, MSU-Net \cite{su2021msu}, Resunet++ \cite{jha2019resunet++} and UcUNet \cite{yang2023ucunet} effectively capture local information from nearby pixels, allowing for fine detail extraction to segment area of interest. However, these methods often struggle with robustness. In contrast, transformers-based models - Swin-UNet \cite{cao2022swin}, Transunet \cite{chen2021transunet}, TransUNet+ \cite{liu2022transunet+} and UGCANet \cite{hung2023ugcanet}, excel at capturing global details through attention mechanisms, making them more robust and generalizable to unseen data. However, attention mechanisms are challenging to train due to their quadratic complexity \cite{han2022survey}.

The hybrid models that combine CNNs and transformer architecture like PolySeg Plus \cite{saad2023polyseg}, UPolySeg \cite{mohapatra2022upolyseg}, RAPUNet \cite{lee2024metaformer}, RaBiT \cite{thuan2023rabit}, and EMCAD \cite{rahman2024emcad} show improvements in results, enhance feature extraction and global contextual understanding. 
DUCK-Net \cite{dumitru2023using} is a U-shaped network having an encode-decoder structure and custom convolutional block designed to process images, capture low-level and high-level features, and demonstrate robustness and generalizability. 
However, challenges remain the same due to variations in polyps shape, low contrast with neighboring tissues, and the occurrence of image abnormalities.

This paper proposed a CLFSeg method that includes an encoder-decoder architecture with different convolution-based blocks and fuzzy modules to make the boundary fuzzier and to tackle the uncertainty and ambiguous data for robustness and precise segmentation.
The CLFSeg model addresses key constraints in biomedical image segmentation by achieving performance comparable to self-attention methods while notably decreasing computing complexity by approximately 30\%. This discourages crisp mask predictions, boosting adaptability in the segmented masks to more effectively address ambiguous and low-contrast areas and facilitates both single-class and multi-class predictions. This paper presents advanced research on CNNs that achieve remarkable performance.
Our contributions are as follows:
\begin{itemize}
\sloppy
\item We propose a CLFSeg model, which consists of several convolutional and fuzzy modules to extract local and global features with fuzzy boundaries for precise, robust segmentation with computational efficiency on the multi-domain dataset.

\item We propose a Fuzzy-Convolutional Module (FCM) consisting of parallel Resnet, Midscope, Widescope, Separable, and Fuzzy Module, then ConvGLU to extract local and global information for noise and uncertainty. 


\item Extensive experiments and ablation are performed to validate that our method achieves remarkable performance on four publicly available benchmark datasets.

\end{itemize}


\section{Related Work}
\label{sec:related_Work}



Classical CNN architectures such as U-Net \cite{ronneberger2015u} and its variants (e.g., U-Net++ \cite{zhou2018unet++}, ResUNet \cite{xiao2018weighted}, and ResUNet++ \cite{jha2019resunet++}) have been the foundation for many polyp and cardiac segmentation methods. These models effectively capture local features via encoder-decoder structures with skip connections, preserving spatial details necessary for fine-grained segmentation. Enhancements like HarDNet-DFUS \cite{liao2022hardnet} improved the backbone and decoder, UPolySeg \cite{mohapatra2022upolyseg} incorporates dilated convolutions, and PolySeg Plus \cite{saad2023polyseg} implements efficient active learning to improve segmentation performance in scenarios with limited labeled data to better handle the diversity of polyp appearances and sizes.
The Duck-Net \cite{dumitru2023using} framework extended U-Net with customized convolutional blocks and residual downsampling, providing multi-resolution features and improving robustness to polyp variability. Similarly, LSSNet \cite{wang2024lssnet} addressed semantic gaps between layers by supplementing local and shallow features, effectively preserving multi-scale information to handle small and fuzzy polyps.
Attention-based models \cite{oukdach2025incolotransnet, xiao2025attention,Kurmi_2019_CVPR} have gained traction for their ability to capture long-range dependencies and focus on salient features. 
PraNet \cite{fan2020pranet} uses reverse attention, ColonFormer \cite{duc2022colonformer}, DUSFormer \cite{10387670}, and SwinPA-Net \cite{9895210} employ Swin Transformer variants to extract global context with different modules, to precisely refine the polyp boundaries with noise suppression.
Polyp-Mamba \cite{xu2024polyp} utilizes state space models to extract long-range dependencies with reduced computational complexity.
QueryNet \cite{Cha_QueryNet_MICCAI2024} brings together polyp segmentation and detection into a single model, and ADSNet \cite{Nguyen_2023_BMVC} introduces a complementary trilateral decoder and continuous attention modules to recover weak features and refine uncertain semantic areas.
Recent work also explores frequency domain integration \cite{10582399, tang2025frequency} and hybrid architectures to enhance segmentation quality. 
MetaFormer \cite{lee2024metaformer} and MSRF-Net \cite{srivastava2021msrf} blend convolutional and transformer modules to achieve a balance between local and global feature extraction at multiple scales. 
RaBiT \cite{thuan2023rabit} adopts a bidirectional feature pyramid network with reverse attention, optimizing the segmentation of polyp borders by refining feature maps across different scales.

Addressing the ambiguity and uncertainty inherent in medical images, especially at boundaries, fuzzy logic has been introduced into segmentation models. FuzzyNet \cite{patel2022fuzzynet} and FuzzyTransNet \cite{liu2023fuzzy} combine fuzzy attention mechanisms with CNNs and transformers to better handle uncertain and noisy features. FANN \cite{nan2023fuzzy} incorporates fuzzy membership functions within attention layers to efficiently manage ambiguous regions. Inspired by biological vision, TransNeXt \cite{shi2024transnext} blends pixel-focused global context with convolutional gating to improve robustness and sensitivity to fine details.



\section{Methodology}
\label{sec:method}

The proposed CLFSeg model illustrated in Figure \ref{fig:CLFSeg_Network}, built on the baseline architecture \cite{dumitru2023using}, with key modifications like incorporating the Fuzzy-Convolution (FC) module as Figure \ref{fig:FC_module} and reducing the complexity of the ResNet block. 
The FC module has a fuzzy logic module and ConvGLU for handling uncertainty and the ambiguous boundary of the polyp and cardiac structures while lowering computational complexity.
Additionally, we provide interpretability using Grad-CAM++ \cite{chattopadhay2018grad}, demonstrating the effectiveness of our methodology. 

In the following subsection, we provide a detailed explanation of the logic and implementation of the FC module, emphasizing its relevance and benefits for medical image segmentation. Furthermore, we outline the approach used to significantly reduce computational complexity while maintaining network performance.


\subsection{CLFSeg Model}
Our model follows an encoder-decoder structure with skip and downscaling connections, along with an FC module. The encoder extracts features and converts them into intermediate representations, while the decoder combines the encoder's intermediate representations to generate the segmentation mask using skip connections. Downscaling connections transfer context to high-resolution layers \cite{ronneberger2015u}, and the FC module, present in both the encoder and decoder, enables feature extraction via five parallel connections. These include:


\begin{figure}[t]


        \begin{center}
        \includegraphics[width=0.8\linewidth]{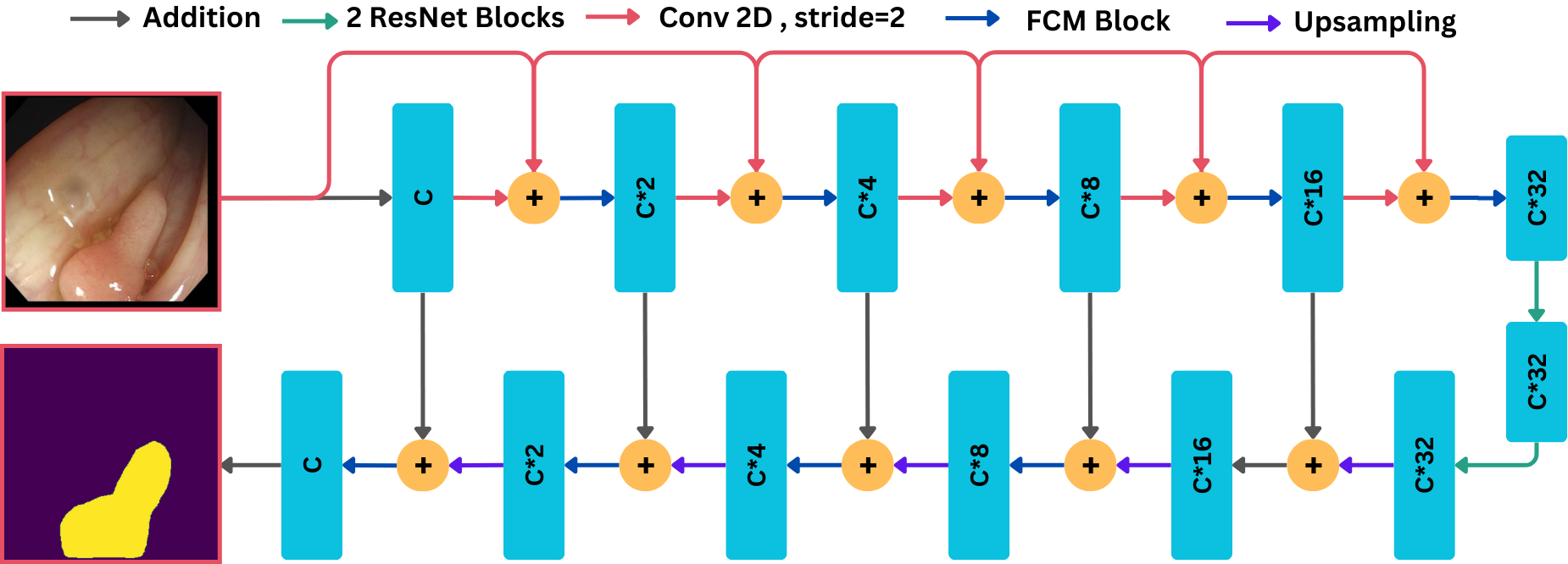}
        \end{center}
        \caption{\small Overview of the CLFSeg model, showing the encoder-decoder structure with skip connections and downscaling layers. It highlights the integration of ResNet blocks, convolutional layers, and the Fuzzy-Convolution (FC) module designed to handle uncertainty and ambiguity in medical image segmentation, especially for polyp and cardiac structures. (Best view in color).}
        \label{fig:CLFSeg_Network}

\end{figure}

\begin{figure}[t]
    

        \begin{center}
        \includegraphics[width=0.6\linewidth]{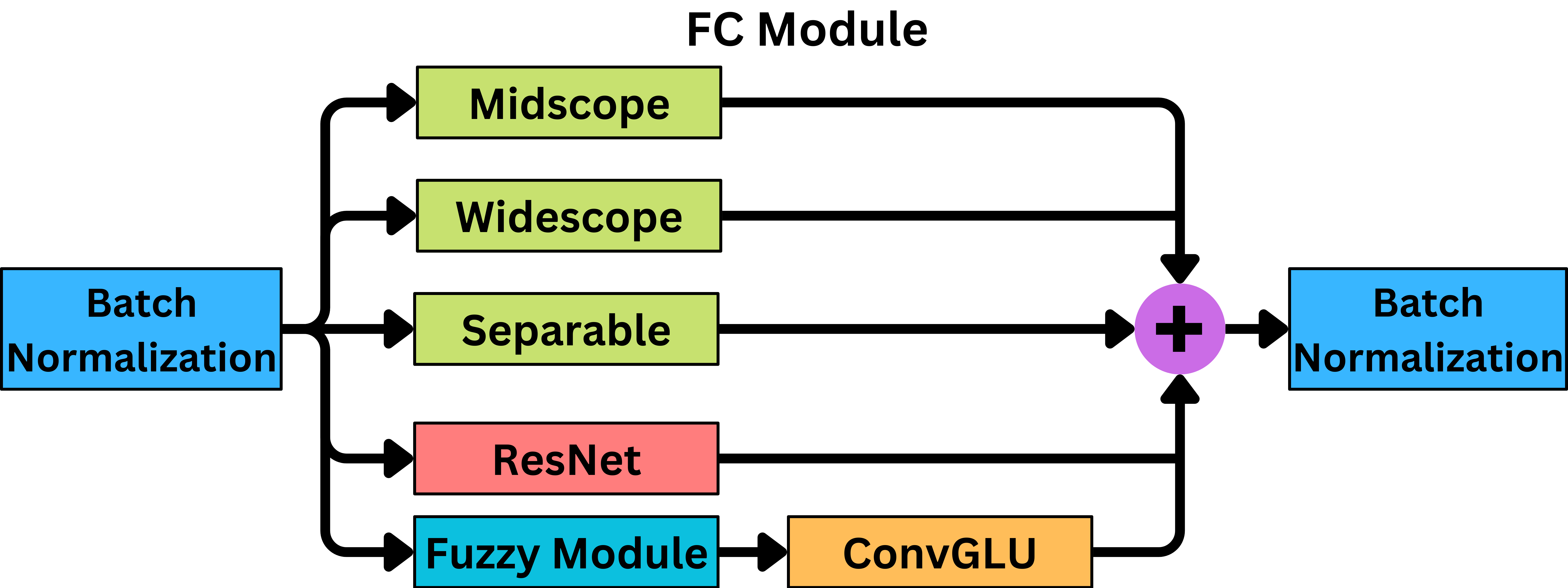}
        \end{center}
        \caption{\small  Overview of FC Module within the CLFSeg architecture, composed of five parallel branches: Midscope, Widescope, Separable, ResNet, and Fuzzy Module, followed by ConvGLU and batch normalization layers. (Best view in color).}
        \label{fig:FC_module}
\end{figure}

\noindent \textbf{1. Midscope block: }
The Midscope block mainly consists of two sequential convolution layers: the first employs a $3 \times 3$ kernel, while the second simulates a $7\times7$ kernel by using a dilation parameter of $2$. This approach allows for capturing a larger global context with reduced computational cost.

\noindent \textbf{2. Widescope block: }Similar to Midscope,  Widescope block applies three consecutive convolutions with a $3 \times 3$ kernel each and a dilation rate of $1, 2$, and $3$, respectively. This configuration simulates kernel sizes of $3 \times 3$, $7 \times 7$, and $15 \times 15$.

\noindent \textbf{3. Separable block:} 
\label{sec:separable}
This block, which simulates the largest kernel, uses $1\times N$ and $N \times 1$ convolutions to mimic an $N \times N$ kernel. This results in the largest local feature representation but may lead to the loss of diagonal features, which can be useful for detecting certain patterns essential for feature extraction.

\noindent \textbf{4. ResNet Block:} 
\label{sec:resnet}
The baseline model \cite{dumitru2023using} employs 6 ResNet layers arranged in 3 parallel connections: the first with a single ResNet block, the second with two serial ResNet blocks, and the third with three serial ResNet blocks. This configuration reportedly enhanced segmentation accuracy by enabling the model to capture finer details. However, in our experiments, we modified this setup to include only a single ResNet connection path, which resulted in: 
a) \textbf{Improved prediction masks}: This improvement is attributed to enhanced gradient propagation and better gradient convergence with a larger area that can be segmented out. 
b) \textbf{Significant reduction in computation}: We observed a reduction of approximately \textbf{30\%} in FLOPs across various filter sizes as viewed in Table \ref{table:CLFSeg_ACDC_Results}.

\noindent \textbf{5. ConvGLU:}
\label{sec:convglu}
The ConvGLU \cite{shi2024transnext} layer, is designed to enhance feature learning capabilities. It consists of dual linear projections and a final projection after feature concatenation. The layer introduces a refined gating mechanism that allows each channel to be influenced by neighboring channels through depthwise convolution. This gating mechanism refines coarse-grained features and suppresses unwanted excitations.

\begin{algorithm}[t]
    \small
    \caption{ \small Fuzzy Module Algorithm}
    \label{alg:FuzzyAttention}
    \begin{algorithmic}[1]
        \REQUIRE Input tensor $x \in \mathbb{R}^{B \times H \times W \times C}$, number of fuzzy sets $n$
        \ENSURE Output tensor after fuzzy module processing
        \STATE Initialize trainable parameters: $w \in \mathbb{R}^{H \times W \times C}$, $b \in \mathbb{R}^{H \times W \times C}$, $\mu \in \mathbb{R}^{1 \times 1 \times 1 \times n \times 1}$, and $\sigma \in \mathbb{R}^{1 \times 1 \times 1 \times n \times 1}$
        
        \STATE Apply element-wise multiplication to the input as
        $x \leftarrow w \odot x + b$
        
        \STATE Perform layer normalization on $x$ as  $ x \leftarrow {LayerNormalization}(x)$

        \STATE Apply LeakyReLU activation function as
        $x \leftarrow {LeakyReLU}(x)$
        
        \STATE Reshape the tensor to $x \in \mathbb{R}^{B \times H \times W \times n \times C}$, where n is the number of fuzzy sets
        
        \STATE Apply Gaussian membership function for each fuzzy set as
        $x \leftarrow \exp\left(-\frac{(x - \mu)^2}{2\sigma^2}\right)$
    
        \STATE Compute the mean across fuzzy set dimension $n$ as
            $x \leftarrow \frac{1}{n}\sum_{i=1}^n x_i \quad \in \mathbb{R}^{B \times H \times W \times C}$
        
        \RETURN $x$
    \end{algorithmic}
\end{algorithm}

\noindent \textbf{6. Fuzzy Module:}
\label{sec:fuzzy}
The Fuzzy Module enhances the model's ability to manage uncertainty and ambiguity in data, particularly in complex tasks like medical image segmentation, where features and boundaries are often unclear \cite{zhang2024fuzzy}. Traditional attention mechanisms and activation functions, such as $sigmoid$, $tanh$, and $LeakyReLU$, struggle with ambiguous data and applying nuanced focus across feature channels. Due to these shortcomings, the Fuzzy Module is introduced within the model. The module combines fuzzy logic with attention mechanisms, enabling the network to assign channel-specific attention through learnable Gaussian membership functions \cite{nan2023fuzzy}. This approach reduces uncertainty in feature representations and improves the network’s focus on relevant features. Additionally, it addresses monotonicity by accounting for a range of variations within data, considering both positive and negative representations. \\
Our objective was to effectively reduce uncertainty by generating fuzzy representations of the input image, resulting in a robust segmentation mask as summarized in Algorithm \ref{alg:FuzzyAttention}. For this, our method uses the Gaussian membership function (Equation \ref{gaussmf}). By working similarly to a weighted average Gaussian filter, it reduces image noise by assigning a degree of membership to each feature. Furthermore, the mean and variance of parameters defined as $\mu$ and $\sigma$, respectively, are trainable, allowing the model to extract features of varying importance.
These fuzzy sets are then summed together, capturing complex patterns and relationships within the data and averaging these membership functions (Equation \ref{gaussmf-mean}), which helps in summarizing the overall values into a single channel. Here, $C$ is the number of fuzzy sets  present in the input image $x$.
\begin{equation}\label{gaussmf}
\scalebox{1}{$
    f(x, \mu , \sigma ) = e^{\frac{-(x - \mu)^{2}}{2\sigma^{2}}}
$}
\end{equation}
\begin{equation}\label{gaussmf-mean}
    \scalebox{1}{$
    \overline{f}(x, \mu , \sigma ) = \frac{1}{C}\sum_{i=0}^{c}e^{\frac{-(x_{i} - \mu_{i})^{2}}{2\sigma_{i}^{2}}}
    $}
\end{equation}

\sloppy
\noindent \textbf{7. Fuzzy-ConvGLU Module:}
Incorporating the ConvGLU layer after the Fuzzy Module  (as illustrated in Figure  \ref{fig:FC_module}) in a neural network architecture significantly enhances its feature-learning capabilities. The Fuzzy Module captures and encodes complex relationships from the input features, producing a rich feature map. When the ConvGLU layer processes this output, the model's ability to extract meaningful information from these detailed features is further enhanced.

Although the ConvGLU layer utilizes a gating mechanism, in the context of our 2D image dataset, it primarily acts as a transformation gating process. The output from the Fuzzy Module, which consists of a single channel, is fed into this layer. ConvGLU then focuses on spatial filtering and gating, allowing it to perform more precise spatial feature selection even with limited channels.
Additionally, the gating mechanism in the ConvGLU layer controls the flow of processed features, ensuring that only the most relevant information is passed forward. This helps the model avoid learning overly generalized features, preserving its sensitivity to finer details in the input data. Each position or token receives a unique gating signal based on local features, helping prevent the model from becoming too coarse in its feature interpretation. 
To support our explanations, we have done an extensive ablation study, as shown in Table~\ref{table:ablation_layers}. Additionally, as demonstrated in Table~\ref{table:CLFSeg_ACDC_Results}, this configuration significantly reduces computational complexity while offering a more effective attention mechanism.

\begin{table}[!ht]
    \begin{center}
     
     \scalebox{0.7}{ 
      \begin{tabular}{c|c|ccccc}
        \toprule
        {Dataset} & {Methods} & { DSC ($\uparrow$)} & {IoU ($\uparrow$)} & {Precision } & {Recall} & {Accuracy}  \\
        \midrule
        \multirow{12}{*}{CVC-ColonDB}
            &FuzzyTransNet \cite{liu2023fuzzy} & 0.7780 & 0.6890 & - & - & - \\
            & UNet \cite{ronneberger2015u} & 0.8032 & 0.7037  & 0.8100 & 0.8274 & 0.9807 \\
            & FuzzyNet (PVT) \cite{patel2022fuzzynet} & 0.8110 & 0.7280 & - & - & - \\
            & PSTNet \cite{10582399} & 0.8270 & 0.7480 & - & - & - \\
            & QueryNet \cite{Cha_QueryNet_MICCAI2024}    & 0.8278 & 0.7593 & 0.8351 & 0.8526 &  - \\
            & LSSNet \cite{wang2024lssnet} & 0.8937 & 0.8221 & - & - & - \\
            & PraNet \cite{fan2020pranet} & 0.9131 & 0.8401  & \textbf{0.9657} & 0.8659 & 0.9901  \\
            & InCoLoTransNet  \cite{oukdach2025incolotransnet}  &  0.9309  &  0.9013  &  0.9177  &  0.9156  & - \\
            & ASRDNet  \cite{xiao2025attention} & 0.9337 & 0.8756 & 0.9327 & 0.9347 & - \\
            & DUCKNet (17 filter) \cite{dumitru2023using}  & 0.9353 & 0.8785 & 0.9314 & 0.9392 & 0.9929  \\
            & DUCKNet (34 filter) \cite{dumitru2023using}  & 0.9230 & 0.8571  & 0.9113 & 0.9351 & 0.9914  \\
            & DUCKNet (24 filter, our-training)  \cite{dumitru2023using} & 0.9389 & 0.8848  & 0.9365 & 0.9413 & 0.9933 \\
            \cmidrule(r){2-7} 
          & \textbf{ CLFSeg (17 filters, OURS)} & 0.9460 & 0.8976   & 0.9582 & 0.9342 & 0.9941 \\
           &  \textbf{CLFSeg (24 filters, OURS)} & 0.9503 & 0.9053 & 0.9583 & \textbf{0.9430} & 0.9943 \\
           &  \textbf{CLFSeg (34 filters, OURS)} & \textbf{0.9593} & \textbf{0.9218}  & 0.9634 & 0.9401 & \textbf{0.9945} \\

         \midrule
            \multirow{14}{*}{CVC-ClinicDB}&
            U-Net \cite{ronneberger2015u} & 0.7631 & 0.6169  & 0.7989 & 0.7303 & 0.9599  \\
            & FAENet \cite{tang2025frequency} & 0.9330 & 0.8830 & - & - & - \\
            & DUSFormer-L \cite{10387670}  &  0.9350  & 0.9020 & 0.9610 & - & - \\
            & FuzzyNet (PVT) \cite{patel2022fuzzynet} & 0.9370 & 0.8890 & - & - & - \\
            &  ADSNet \cite{Nguyen_2023_BMVC}   & 0.9380 & 0.8900 & - & - & - \\     
            & SwinPA-Net \cite{9895210} & 0.9410 & 0.8940 & - & - & - \\
            & FuzzyTransNet \cite{liu2023fuzzy} & 0.9420 & 0.8910 & - & - & - \\
            & PSTNet \cite{10582399}  & 0.9450 & 0.9010 & - & - & - \\
            & CTHP \cite{xue2024lighter} & 0.9471 & 0.9021 & 0.9524 & 0.9444 & - \\
            & Duck-Net (17 filters) \cite{dumitru2023using} & 0.9450 & 0.8952& 0.9488 & 0.9406 & 0.9903  \\
            & Duck-Net (34 filters) \cite{dumitru2023using} & 0.9478 & 0.9009 & 0.9468 & \textbf{0.9489} & 0.9907  \\
            & Duck-Net (24 filters, our-training) \cite{dumitru2023using} & 0.9430 & 0.8922 & 0.9539 & 0.9324 & 0.9900 \\
            & Polyp-Mamba  \cite{xu2024polyp}  & 0.9490 & 9070 & - & - & - \\
            \cmidrule(r){2-7} 
            & \textbf{CLFSeg (17 filters, OURS)} & \textbf{0.9533} & \textbf{0.9108} & \textbf{0.9636} & 0.9432 & \textbf{0.9918}  \\
            & \textbf{CLFSeg (24 filters, OURS)} & 0.9530 & 0.9103 & 0.9594  & 0.9467 & 0.9917 \\
            & \textbf{CLFSeg (34 filters, OURS)} & 0.9500 & 0.9048 & 0.9568 & 0.9433 & 0.9912  \\

        \midrule
         \multirow{12}{*}{ETIS-LaribPolyp} 
            & FuzzyNet (PVT) \cite{patel2022fuzzynet} & 0.7910 & 0.7020 & - & - & - \\
            & U-Net \cite{ronneberger2015u} & 0.7984 & 0.6969 &   0.8322 &  0.7724  & 0.9734  \\
            & PSTNet \cite{10582399}  & 0.8000 & 0.7260 & - & - & - \\
            & QueryNet \cite{Cha_QueryNet_MICCAI2024} & 0.8189 & 0.7399 & 0.7488 & 0.7740 & - \\
             & Polyp-Mamba  \cite{xu2024polyp}  & 0.8250 & 0.7470 & - & - & - \\
            & PraNet \cite{fan2020pranet} & 0.8827 & 0.7900 &  0.9825  &  0.8013  & 0.9877  \\
            & FCN-Transformer \cite{sanderson2022fcn} & 0.9163 & 0.8455 &  \textbf{0.9633}  & 0.8736   & 0.9915  \\
            &  ASRDNet \cite{xiao2025attention} & 0.9313  &  0.8714   &  0.9055  & 0.9586 & - \\
            & InCoLoTransNet  \cite{oukdach2025incolotransnet}   &  0.9316  & 0.9225  & 0.9283  & 0.9392  & - \\
            & Duck-Net (17 filters) \cite{dumitru2023using} & 0.9324 & 0.8734 &  0.9539  &  0.9118  &  0.9930 \\
            & Duck-Net (34 filters) \cite{dumitru2023using} & 0.9354 & 0.8788 &  0.9309  &  0.9400  & 0.9931  \\
            & Duck-Net (24 filters, our-training) \cite{dumitru2023using} & 0.9396 & 0.8861  & 0.9372 & 0.8861 & 0.9936 \\
            \cmidrule(r){2-7} 
            & \textbf{CLFSeg (17 filters, OURS)} & 0.9292 & 0.8678 & 0.9189 & 0.9396 & 0.9924 \\
            & \textbf{CLFSeg (24 filters, OURS)} & 0.9140 & 0.8416 &  0.8826  &  \textbf{0.9476} & 0.9905 \\
            & \textbf{CLFSeg (34 filters, OURS)} & \textbf{0.9487} & \textbf{0.9024}   & 0.9596 & 0.9380 & \textbf{0.9946} \\
            
        \bottomrule
      \end{tabular}
      }
    \end{center}
    \caption{ \small  Comparing the performance of different segmentation methods on the CVC-ColonDB, CVC-ClinicDB, and ETIS-LaribPoly dataset. Bold shows best model results.}
     \label{table:CLFSeg_CVC-ColonDB_CVC-ClinicDB_ETIS-LaribPolyp_Results}
\end{table}

\begin{table}[!ht]
    \begin{center}
    
    \scalebox{0.8}{
      \begin{tabular}{c|cc|cc}
        \toprule
     {Methods} & { DSC ($\uparrow$)} & {IoU ($\uparrow$)} & {\#FLOPs (M) $\downarrow$} & { \#Params (M) $\downarrow$} \\
        \midrule
           U-Net \cite{ronneberger2015u} & 0.8755    & - & - &  - \\
            EM-ViT \cite{karimijarbigloo2025frequency}  & 0.9029 & - & - & - \\
            CSWin-UNet \cite{liu2025cswin}  & 0.9146 & - & - & - \\
            Lite-MixedNet \cite{ren2025lite} & 0.9207 & - & - & - \\  
            PVT-EMCAD-B2 \cite{rahman2024emcad} &  0.9212  &    \\
            RWKV-UNet \cite{su2025vmkla} & 0.9217 & - & - & - \\
            MIST \cite{rahman2024mist} &  0.9256  & - & - &  -   \\
            LHU-Net \cite{sadegheih2024lhu}  &  0.9266  & - & - &  -   \\
            FCT \cite{tragakis2023fully} &  0.9302  & - & - &  -  \\
            Adaptive t-vMF \cite{kato2022adaptive} &  0.9368  & 0.8851 & - &  -   \\


        \midrule
           Duck-Net (17 filters, our-training) \cite{dumitru2023using} &  0.9392  &  0.8853   & 38  & 38   \\
            Duck-Net (24 filters, our-training)  \cite{dumitru2023using} &  0.9470 &  0.8993  & 77  & 77 \\
           Duck-Net (34 filters, our-training) \cite{dumitru2023using} &  0.9428  &  0.8917    & 155 &  155  \\
        \midrule
           \textbf{CLFSeg (17 filters, OURS)} &  \textbf{0.9522} & \textbf{0.9087}   & \textbf{26}  & \textbf{38} \\
           \textbf{CLFSeg (24 filters, OURS)} &  0.9403  & 0.8873  & \textbf{54}  & \textbf{71} \\
            \textbf{CLFSeg (34 filters, OURS)}&  0.9498  & 0.9043   & \textbf{106} & \textbf{132} \\
     
        \bottomrule
      \end{tabular}
    }
    \end{center}
    \caption{ \small  Comparing the performance of different segmentation methods on the ACDC dataset. Bold shows the best model results.}
 \label{table:CLFSeg_ACDC_Results}

\end{table}

\begin{table}[!ht]
    \begin{center}
   
    \scalebox{0.85}{
    \begin{adjustbox}{width=\linewidth}  
        
          \begin{tabular}{ccc|ccccc}
            \toprule
            \multicolumn{3}{c|}{Layers} & \multirow{2}{*}{DSC ($\uparrow$)} & \multirow{2}{*}{IoU ($\uparrow$)} & \multirow{2}{*}{Precision} & \multirow{2}{*}{Recall} & \multirow{2}{*}{Accuracy} \\
            \cmidrule(r){1-3} 
         {Fuzzy Module} & {ConvGLU} & {1-ResNet} & { } & {} & { } & {} & {} \\
            \midrule
                & & & {0.9389} & {0.8848}  & {0.9365} & {0.9413} & {0.9933}\\
              \checkmark  &     &     &  0.9374  &  0.8823  &  0.9279  &  0.9472  &  0.9930  \\
                 &    \checkmark  &     &  0.9337  &  0.8757  & \textbf{0.9657}   &  0.9287  &  0.9943  \\
                 &     &   \checkmark   & 0.9269   &  0.8638  &  0.9489  &  0.9059  &  0.9921  \\
               \checkmark  &   \checkmark   &     &  0.9469  &  0.8991  &  0.9476  &  0.9202  & 0.9928   \\
                 \checkmark  &     &  \checkmark   &  0.9491  & 0.9031   &  0.9517  &  \textbf{0.9520} &  \textbf{0.9947}  \\
                &    \checkmark  &    \checkmark  &  0.9277  &  0.8651  &  0.9627  & 0.8951  &  0.9923  \\
                 \checkmark &   \checkmark   &    \checkmark  &  \textbf{0.9503}  &  \textbf{0.9053}  &  0.9583  &  0.9430  &  0.9943  \\
         
            \bottomrule
          \end{tabular}
    
        \end{adjustbox}
        }
    \end{center}
     \caption{ \small  Ablation study based on different layers (Fuzzy module, ConGLU, and ResNet) on CVC-ColonDB dataset with 24 filters. Bold highlights the best results.}
    \label{table:ablation_layers}
\end{table}

\begin{table}[t]
    \begin{center}
    
    \scalebox{0.85}{
    \begin{adjustbox}{width=\linewidth}  
        
          \begin{tabular}{c|ccc|ccccc}
            \toprule
           \multirow{2}{*}{Filters} & \multicolumn{3}{c|}{Layers} & \multirow{2}{*}{DSC ($\uparrow$)} & \multirow{2}{*}{IoU ($\uparrow$)} & \multirow{2}{*}{Precision} & \multirow{2}{*}{Recall} & \multirow{2}{*}{Accuracy} \\
            \cmidrule(r){2-4} 
               &    {1-resnet} & {2-resnet} & {3-resnet} & { } & {} & { } & {} & {} \\
            \midrule
            
             \multirow{3}{*}{17 Filters}  &     \checkmark  &     &     &  \textbf{0.9460}  & \textbf{0.8976}   &  \textbf{0.9582}  &  0.9342  & \textbf{0.9941}   \\
                &   \checkmark  &    \checkmark  &     &  0.9416  &  0.8897  &  0.9402  &  \textbf{0.9430}  &  0.9936  \\
                 &  \checkmark &  \checkmark   &   \checkmark   &  0.8799  &  0.7857  &  0.8507  &  0.9113  &  0.9864  \\
    
            \midrule
                \multirow{3}{*}{24 Filters}  &     \checkmark  &     &     &  \textbf{0.9503}  &  \textbf{0.9053}  &  0.9583  & \textbf{ 0.9430}  &  \textbf{0.9943}  \\
                &   \checkmark    &    \checkmark  &     &  0.9451  &  0.8959  &  0.9522  &  0.9380  & 0.9940   \\
                 & \checkmark  &   \checkmark  &   \checkmark   &  0.9448  &  0.8953  &  \textbf{0.9608}  &  0.9293  &  0.9940  \\
    
             \midrule
                \multirow{3}{*}{34 Filters}  &     \checkmark  &     &     &  \textbf{0.9495}  &  \textbf{0.9039}  &  \textbf{0.9624}  &  0.9370  & \textbf{0.9945}   \\
                &   \checkmark  &    \checkmark  &     &  0.9380  &  0.8832  &  0.9599  &  0.9170  &  0.9933  \\
                 &   \checkmark  & \checkmark  &   \checkmark   &  0.9455  & 0.8966   &  0.9422  &  \textbf{0.9488}  & 0.9940   \\

            \bottomrule
          \end{tabular}
    
        \end{adjustbox}
    }
    \end{center}
     \caption{ \small  Ablation study based on different parallel ResNet blocks on CVC-ColonDB dataset with 24 filters. Bold highlights the best results.}
    \label{table:ablation_different_resnet_block}

\end{table}

\begin{table}[!ht]
    \begin{center}
    
     \scalebox{0.65}{
    \begin{adjustbox}{width=\linewidth}  
        
          \begin{tabular}{ccc|ccccc}
            \toprule
            \multicolumn{3}{c|}{Losses with 24 filters} & \multirow{2}{*}{DSC ($\uparrow$)} & \multirow{2}{*}{IoU ($\uparrow$)} & \multirow{2}{*}{Precision} & \multirow{2}{*}{Recall} & \multirow{2}{*}{Accuracy} \\
            \cmidrule(r){1-3} 
         {BCE} & {Dice} & {Focal} & { } & {} & { } & {} & {} \\
            \midrule
            
              \checkmark  &     &     &  0.9423  &  0.8909  &  \textbf{0.9623}  & 0.9231   &  0.9938  \\
                 &    \checkmark  &     &  0.9379  &  0.8832  &  0.9525  &  0.9238  &  0.9933  \\
               \checkmark  &   \checkmark   &     &  \textbf{0.9503 } &  \textbf{0.9053 } &  0.9583  &  0.9430  &  \textbf{0.9943}  \\
                &    \checkmark  &    \checkmark  &  0.9455  &  0.8966  &  0.9401  &  \textbf{0.9509}  &  0.9940  \\

            \bottomrule
          \end{tabular}

    \end{adjustbox}
    }
    \end{center}
    \caption{ \small  Ablation study based on different loss functions on CVC-ColonDB dataset with 24 filters. Bold highlights the best results.}
     \label{table:ablation_losses}

\end{table}

\begin{table}[!ht]
    \begin{center}
   \begin{adjustbox}{width=\linewidth}  
     
     \scalebox{0.85}{ 
      \begin{tabular}{c|c|c|c|c|c|c|c|c|c|c|c|c}
        \toprule
        Dataset & \multicolumn{4}{c|}{CVC-ColonDB} & \multicolumn{4}{c|}{CVC-ClinicDB} & \multicolumn{4}{c}{ACDC} \\
        \midrule
        Filters & \multicolumn{2}{c}{17} & \multicolumn{2}{|c|}{34} & \multicolumn{2}{c}{17} & \multicolumn{2}{|c|}{34} & \multicolumn{2}{c}{17} & \multicolumn{2}{|c}{34} \\
        \midrule
        Methods & DuckNet & CLFSeg & DuckNet & CLFSeg & DuckNet & CLFSeg & DuckNet & CLFSeg & DuckNet & CLFSeg & DuckNet & CLFSeg \\
        \midrule
        HD95 & 1.97 & \textbf{1.43} & 2.00 & \textbf{1.50} & 4.44 & \textbf{3.54} & 6.39 & \textbf{5.97} & 0.90 & \textbf{0.58} & 1.36 & \textbf{0.81} \\

        \bottomrule
      \end{tabular}
      }
     \end{adjustbox}
    \end{center}
    \caption{ \small  Analysis of boundary clarity on DuckNet and proposed CLFSeg methods across datasets and filter settings (lower HD95 is better).}
     
     \label{table:ablation_hd95}
\end{table}

\section{Experimental Results}
\label{sec:results}

\textbf{Dataset}: \textit{CVC-ColonDB} consists of 380 images and their corresponding ground truth labels with $574 \times 500$ resolution. \textit{CVC-ClinicDB} consists of 612 images and corresponding pixel-wise annotations with $384 \times 288$ resolution from 31 colonoscopy sequences. \textit{ETIS-LaribPolyp} consists of 196 images and respective ground-truth labels with $1255 \times 966$ resolution. \textit{Automated Cardiac Diagnosis Challenge (ACDC)} is an MRI image dataset containing 100 cases and 3 labels.

\noindent  \textbf{Implementation Setup:} The CVC-ColonDB, ETIS-LaribPolypDB, and CVC-ClinicDB datasets are split into 80:10:10 for training, validation, and test sets, while for ACDC, the ratio is 70:10:20, i.e., 1312:380:210 images. We apply data augmentation using standard flips on the X and Y axes, color jitter, and affine transformations. The model is designed to handle both binary and multi-class segmentation tasks, generating reliable segmentation saliency maps in each case. The training process used a batch size of 4 over 1000 epochs, input image size is 352x352 pixels, the learning rate is 1e-4, and the optimizer is RMSprop \cite{tieleman2012lecture}. The model is implemented with the TensorFlow \cite{abadi2016tensorflow} framework and trained on an NVIDIA A100 GPU.
\noindent We employ a hybrid BCE and dice Loss in equal proportions, and the Dice Similarity Coefficient (DSC), Intersection over Union (IoU), precision, recall, and accuracy are used for evaluation of the segmentation models. This approach encourages the model to achieve pixel-level performance while preserving global region consistency.

\begin{figure}[t]
  \begin{center}
  \scalebox{0.92}{
  \begin{minipage}{1\linewidth}
    \includegraphics[width=0.47\linewidth]{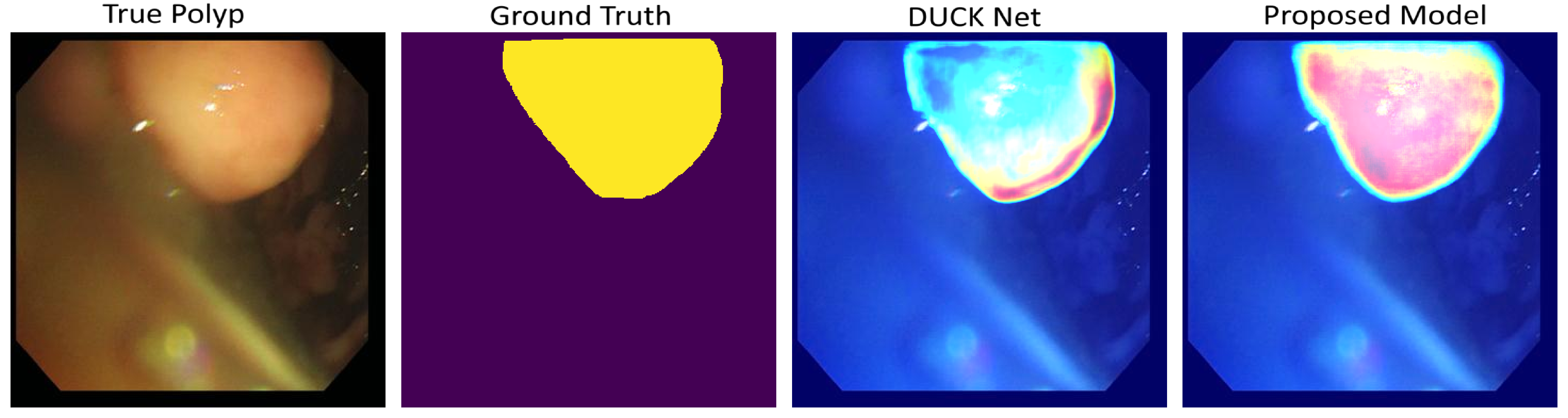}
    \includegraphics[width=0.47\linewidth]{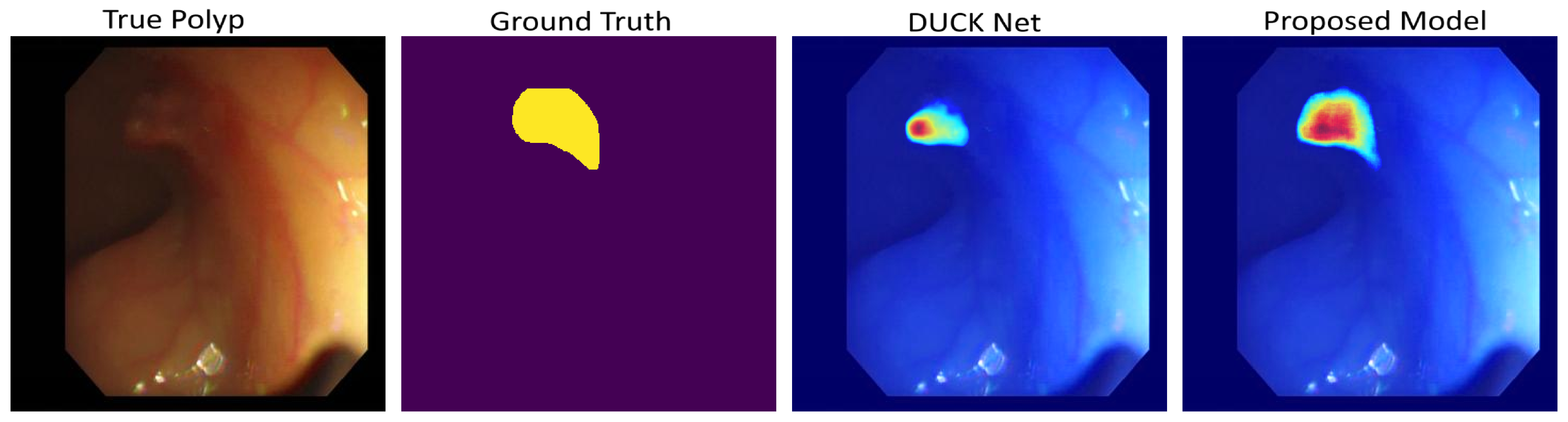}

    \includegraphics[width=0.47\linewidth]{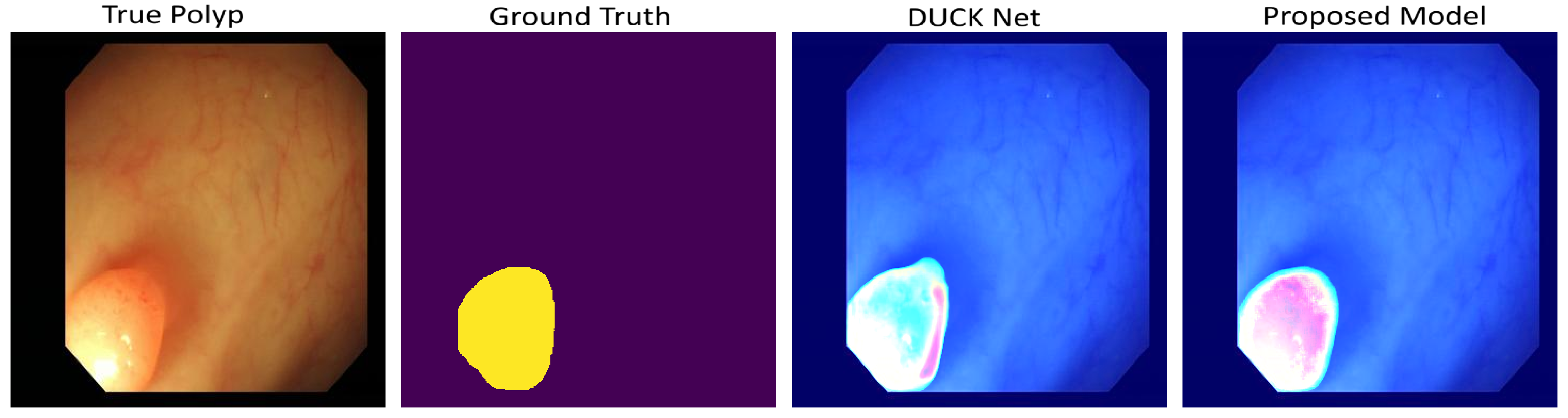}
    \includegraphics[width=0.47\linewidth]{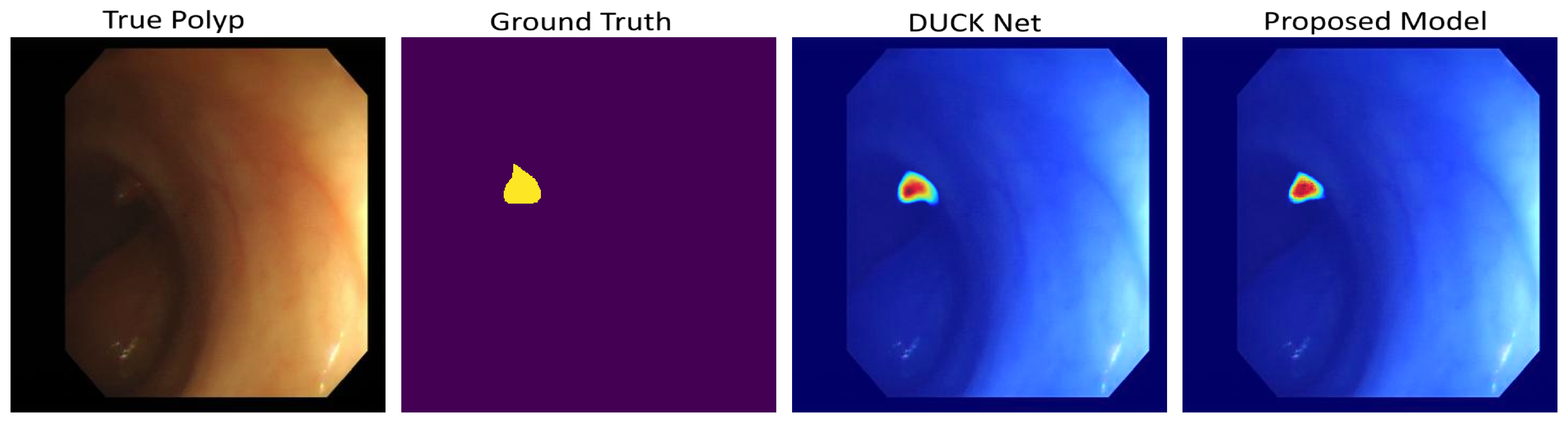}
    \label{}
  \end{minipage}
    }  
    \end{center}
    \caption{ \small Comparison of GradCam++ Visualization between DuckNet and CLFSeg model on CVC-ColonDB dataset. (Best view in color).}
    \label{fig:GradCam_visual}
\end{figure}

\begin{figure}[t]
    \begin{center}
    \scalebox{0.92}{
    \begin{minipage}{1\linewidth}
        \includegraphics[width=0.47\linewidth]{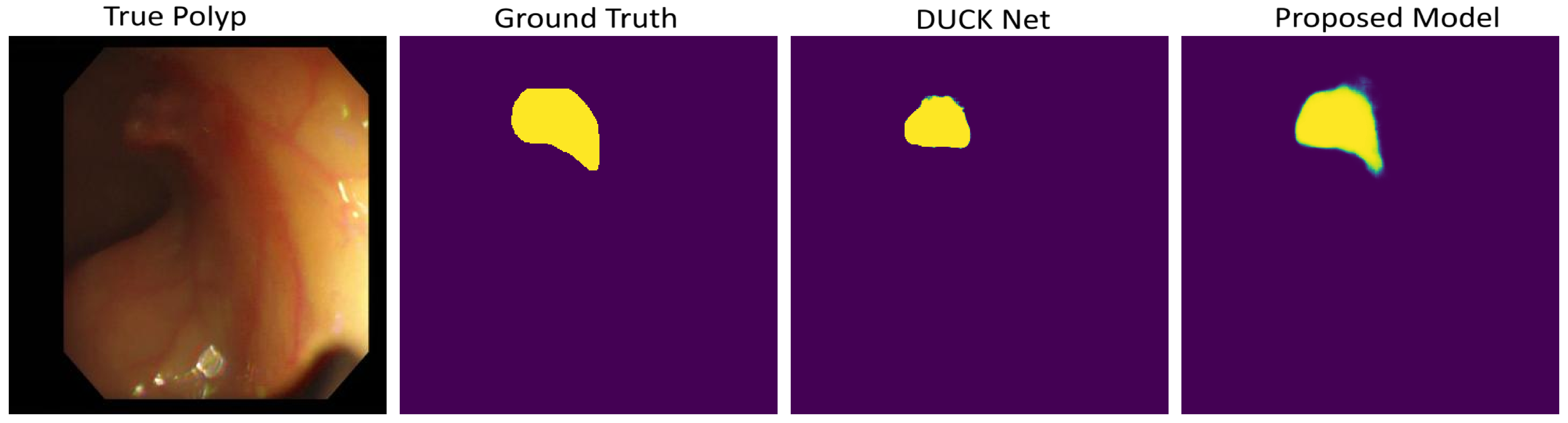}
         \hspace{0.8em}
         \includegraphics[width=0.47\linewidth]{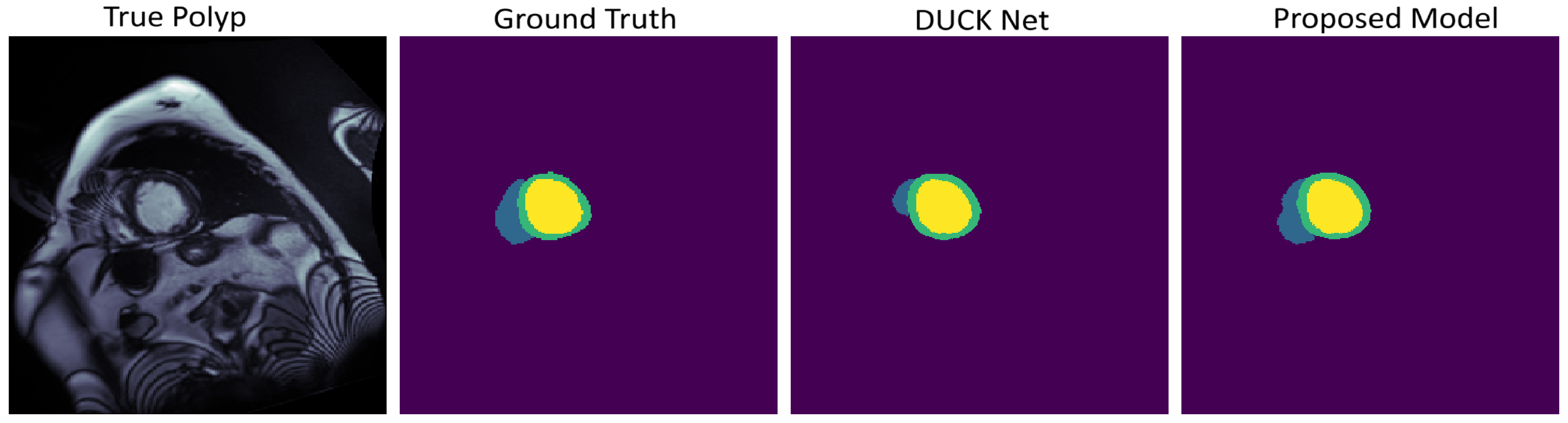}
         
         
         \includegraphics[width=0.47\linewidth]{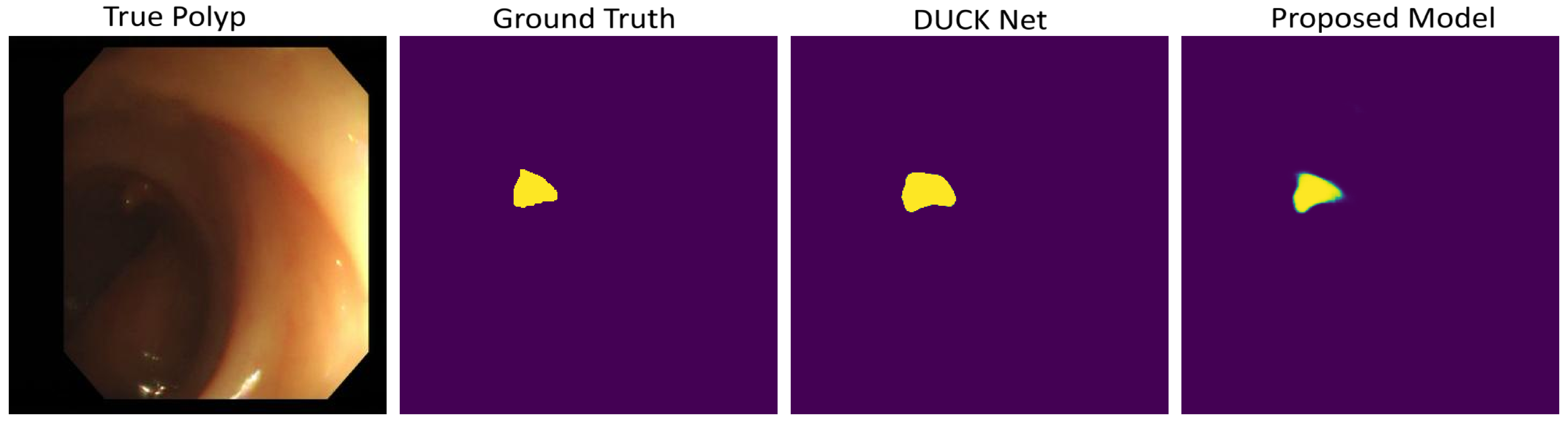}
          \hspace{0.8em}
         \includegraphics[width=0.47\linewidth]{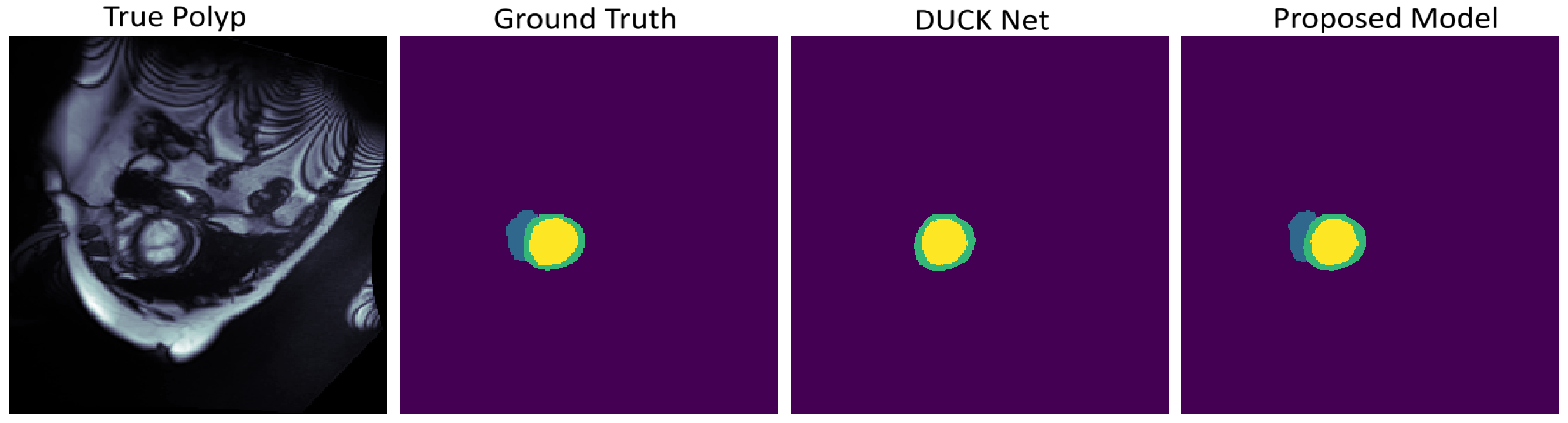}

    \end{minipage}
 }
 \end{center}
    \caption{ \small Comparison of Segmentation Map from proposed CLFSeg model for CVC-ColonDB (left side), and ACDC dataset (right side). (Best view in color).}
    \label{fig:Segmentation_Map_visual}

\end{figure}

\section{Comparison with SOTA}

The proposed CLFSeg model achieves significant improvements over existing state-of-the-art methods on multiple polyp datasets shown in  Table \ref{table:CLFSeg_CVC-ColonDB_CVC-ClinicDB_ETIS-LaribPolyp_Results}. On the CVC-ColonDB polyp dataset, CLFSeg attained a DSC of \textbf{0.9593}, outperforming prior leading methods like DUCKNet (DSC 0.9389), ASRDNet (DSC 0.9337), and PraNet (DSC 0.9131) by more than \textbf{2\%}. Similarly, on the CVC-ClinicDB dataset, CLFSeg reached a DSC of \textbf{0.9533}, surpassing SOTA methods. On the ETIS-LaribPolyp dataset, CLFSeg also demonstrated remarkable gains, achieving a DSC of \textbf{0.9487}, outperforming earlier models by a margin of approximately \textbf{1\%} compared to baseline DuckNet \cite{dumitru2023using} architectures. This performance reflects the proposed CLFSeg's superior capability in capturing both local and global features with fuzzy boundary refinement.

In the cardiac segmentation task on the ACDC dataset, CLFSeg match the performance with baseline \cite{dumitru2023using} for 17 and 34 filters. But exceeded the performance of recent models such as Duck-Net and Adaptive t-vMF for 17 filters, reaching a DSC of \textbf{0.9522} and IoU of \textbf{0.9087} while reducing computational complexity by up to \textbf{30\%} as shown in Table \ref{table:CLFSeg_ACDC_Results}.

\subsection{Ablation Study}

\noindent \textbf{1. Analysis of the Different Layers of the CLFSeg model: }Three layers—Fuzzy Module, ConvGLU, and 1-ResNet are evaluated in the ablation studies, as shown in Table \ref{table:ablation_layers}. With a fixed filter size of 24, these layers underwent extensive testing. Results indicate that the combination of Fuzzy Module with ConvGLU and 1-ResNet achieved optimal segmentation maps, reaching \textbf{0.9503} DSC and \textbf{0.9053} IoU. The findings demonstrate the effectiveness of our proposed approach, showing a \textbf{+2.46\%} improvement over the single use of the 1-ResNet layer. From the Table \ref{table:CLFSeg_ACDC_Results}, we conclude that 1-ResNet used in our proposed CLFSeg model achieves good performance with low computational cost compared to baseline Duck-Net \cite{dumitru2023using}.

\noindent \textbf{2. Analysis of the effect of different filter sizes:} In Table \ref{table:CLFSeg_CVC-ColonDB_CVC-ClinicDB_ETIS-LaribPolyp_Results}, we observe that increasing the filter size improved DSC of CVC-ColonDB dataset from \textbf{0.9460} to \textbf{0.9593}, marking a \textbf{+1.38\%} increase. Our model, with a filter size of 34, delivered SOTA performance on CVC-ColonDB dataset, with configurations of 24 and 17 filters surpassing benchmarks \cite{dumitru2023using}.

\noindent \textbf{3. Analyse parallel ResNet block: }The impact of using parallel ResNet blocks with an increasing number of layers is analyzed, as outlined in Table \ref{table:ablation_different_resnet_block}. Results show that adding more parallel ResNet blocks negatively impacted model efficiency, despite thorough testing with all three filter sizes. A single ResNet block effectively captured fine details, while additional blocks merely extended training time and added complexity, leading to an average \textbf{4.65\%} decrease in performance across all filter sizes when additional blocks are used.

\noindent \textbf{4. Analysis of the boundary Clarity: }
Our evaluation considers HD95, a boundary-aware metric that measures the maximum distance between the true and predicted segmentation boundaries while discarding extreme outliers (95\% confidence interval). Our model consistently achieves superior performance on this metric. Table \ref{table:ablation_hd95} reports the improvements across 17- and 34-filter configurations for CVC-ColonDB, CVC-ClinicDB, and ACDC datasets. On average, CLFSeg reduces the HD95 score by 0.54 units compared to DUCKNet.

\noindent \textbf{5. Analysis of the different loss functions: }In Table \ref{table:ablation_losses}, hybrid BCE-dice loss produced a DSC of \textbf{0.9503}, while BCE loss and dice loss alone resulted in \textbf{0.9423} and \textbf{0.9379}, respectively. The Focal and dice loss achieved a DSC of \textbf{0.9455}, demonstrating that an equal contribution of BCE and dice losses results in the best model performance.


\noindent \textbf{6. Visualization of Activation Maps: }
{Grad-CAM++} provides a visual representation of the regions focused on by the model. Figure \ref{fig:GradCam_visual} shows that CLFSeg more accurately captures the entire ground truth for both small and large regions compared to DuckNet, which often overlooks relevant polyp areas. The red-highlighted regions are more prominent in CLFSeg, indicating its ability to focus on clinically significant areas, particularly due to the FC module. CLFSeg effectively captures finer details while making confident predictions, as evidenced by the red regions in Grad-CAM++. This underscores the model’s precision and interpretability, enabling improved boundary delineation across varying polyp sizes.

%

\noindent \textbf{7. Visualization of Segmentation Map: }Figure \ref{fig:Segmentation_Map_visual} compares \textbf{segmentation maps} across CVC-ColonDB and ACDC datasets. In CVC-ColonDB, the proposed CLFSeg model outperforms DuckNet, where DuckNet misclassifies relevant regions, while CLFSeg accurately highlights polyp boundaries and uncertainty areas. In the ACDC dataset, the proposed CLFSeg effectively segments and differentiates the different cardiac regions. It helps in conditions where a single image consists of multiple areas of interest and precise boundary delineation is important, like cancer treatment.

\section{Conclusion}
\label{sec:conclusion}


The proposed CLFSeg framework enhances automated medical image segmentation tasks with a fast, efficient segmentation mask, highlighting the FC module. Fuzzy logic and a refining layer reduce uncertainty and enable flexible boundary fuzzification for more adaptable segmentation. Achieving state-of-the-art DSC scores of 0.9593, 0.9533, 0.9487, and 0.9522 on CVC-ColonDB, CVC-ClinicDB, ETIS-LaribPolypDB, and ACDC demonstrates high effectiveness. This approach advances AI in healthcare and sets a foundation for future improvements, including broader datasets and privacy-focused solutions.

\bibliography{egbib}
\end{document}